\documentclass[10pt,twocolumn,letterpaper]{article}

\usepackage{cvpr}
\usepackage{times}
\usepackage{epsfig}
\usepackage{graphicx}
\usepackage{amsmath}
\usepackage{amssymb}

\usepackage{times}
\usepackage{epsfig}
\usepackage{graphicx}
\usepackage{amsmath}
\usepackage{amssymb}
\usepackage{xspace}
\usepackage{nicefrac}
\usepackage{subcaption}
\usepackage{soul}
\usepackage{booktabs}       
\usepackage{multirow}

\DeclareFontFamily{OT1}{pzc}{}
\DeclareFontShape{OT1}{pzc}{m}{it}{<-> s * [1.10] pzcmi7t}{}
\DeclareMathAlphabet{\mathpzc}{OT1}{pzc}{m}{it}

\usepackage[breaklinks=true,bookmarks=false]{hyperref}

\cvprfinalcopy 

\def\cvprPaperID{9570} 

\ifcvprfinal\fi
\begin{document}

\title{Learning to Have an Ear for Face Super-Resolution}

\author{Givi Meishvili \quad Simon Jenni \quad Paolo Favaro\\
University of Bern, Switzerland\\
{\tt\small \{givi.meishvili, simon.jenni, paolo.favaro\}@inf.unibe.ch}
}

\maketitle
\thispagestyle{empty}

\begin{abstract}
We propose a novel method to use both audio and a low-resolution image to perform extreme face super-resolution (a $16\times$ increase of the input size). When the resolution of the input image is very low (e.g., $8\times 8$ pixels), the loss of information is so dire that important details of the original identity have been lost and audio can aid the recovery of a plausible high-resolution image. In fact, audio carries information about facial attributes, such as gender and age. 
To combine the aural and visual modalities, we propose a method to first build the latent representations of a face from the lone audio track and then from the lone low-resolution image. 
We then train a network to fuse these two representations. 
We show experimentally that audio can assist in recovering attributes such as the gender, the age and the identity, and thus improve the correctness of the high-resolution image reconstruction process. Our procedure does not make use of human annotation and thus can be easily trained with existing video datasets.
Moreover, we show that our model builds a factorized representation of images and audio as it allows one to mix low-resolution images and audio from different videos and to generate realistic faces with semantically meaningful combinations.
\end{abstract}


\section{Introduction}
Image super-resolution is the task of recovering details of an image that has been captured with a limited resolution. 
Typically, the resolution of the input image is increased by a scaling factor of $4\times$ to $8\times$.
\begin{figure}[t!]
    \centering
    		\includegraphics[width=\linewidth]{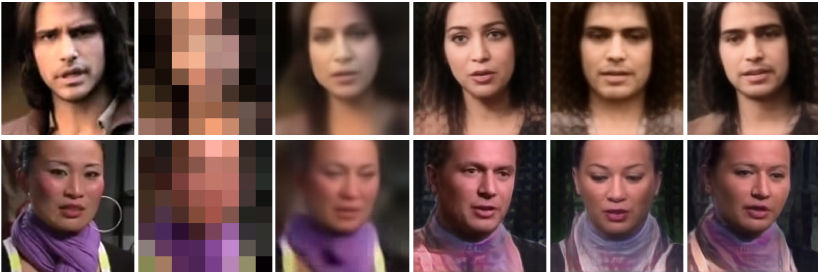}\\
		(a)\hspace{.12\linewidth}(b)\hspace{.12\linewidth}(c)\hspace{.12\linewidth}(d)\hspace{.12\linewidth}(e)\hspace{.12\linewidth}(f)
    \caption{\textbf{Audio helps image super-resolution.} (a) and (b) are the ground-truth and $16\times$ downsampled images respectively; (c) results of the SotA super-resolution method of Huang~\etal~\cite{Huang_2017_ICCV}; (d) our super-resolution from only the low-res image; (e) audio only super-resolution; (f) fusion of both the low-res image and audio. In these cases all methods fail to restore the correct gender without audio.}
  	\label{fig:teaser}
  	\vspace{-4mm}
\end{figure}
In the more extreme case, where the scaling factor is $16\times$ or above, the loss of detail can be so considerable that important semantic information is lost. This is the case, for example, of images of faces at an $8\times 8$ pixels resolution, where information about the original identity of the person is no longer discernible. The information still available in such a low-resolution image is perhaps the viewpoint and colors of the face and the background. While it is possible to hallucinate plausible high-resolution images from such limited information, useful attributes such as the identity or even just the gender or the age might be incorrect (see Fig.~\ref{fig:teaser}~(a)-(d)).

\begin{figure}[t!]
    \centering
        \includegraphics[width=.99\linewidth,trim=1.3cm 2.8cm 3cm .5cm, clip]{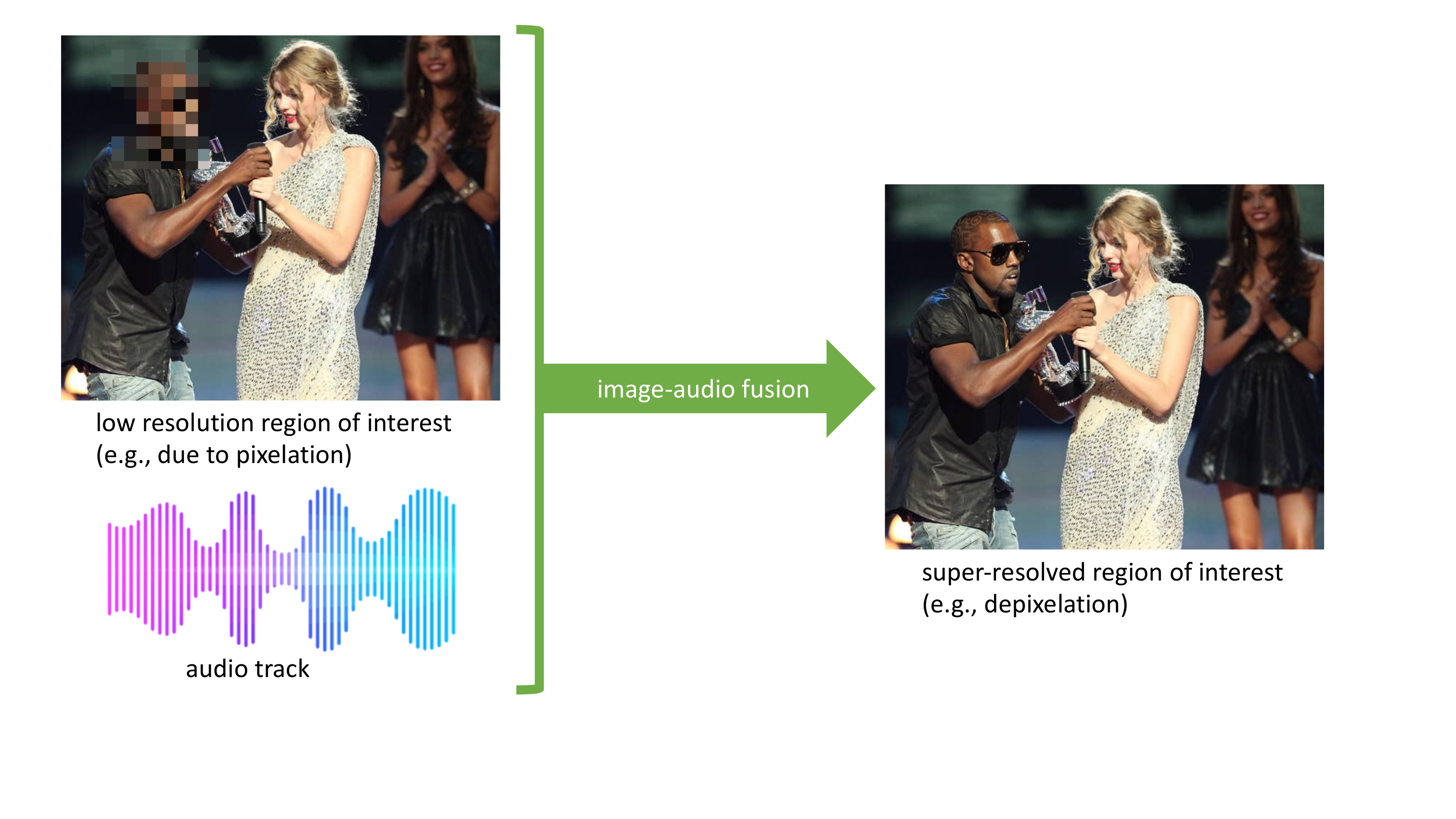}
    \caption{Pixelation is used to hide the identity of a person (left). However, audio could assist in recovering a super-resolved plausible face (right).}
  	\label{fig:real_mixing}
  	\vspace{-4mm}
\end{figure}

If the low-resolution image of a face is extracted from a video, we could also have access to the audio of that person. Despite the very different nature of aural and visual signals, they both capture some shared attributes of a person and, in particular, her identity. In fact, when we hear the voice of an iconic actor we can often picture his or her face in our mind. 
\cite{oh2019speech2face} recently showed that such capability can be learned by a machine as well.
The possibility to recover a full identity is typically limited to a set of known people (\eg, celebrities). 
Nonetheless, even when the identity of a person is completely new, her voice indicates important facial attributes such as gender, age, and ethnicity. If such information is not present in the visual data (\eg, with a low-resolution image), audio could be a benefit to image processing and, in particular, image super-resolution (see Fig.~\ref{fig:teaser}~(e)-(f)). 
For example, in videos where the identity of a speaker is hidden via pixelation, as shown in Fig.~\ref{fig:real_mixing}, audio could be used to recover a more plausible face than from the lone low-resolution image.

\begin{figure*}[t!]
    \centering
        \includegraphics[width=.9\linewidth,trim=0.5cm 0.5cm 2.2cm 4.2cm, clip]{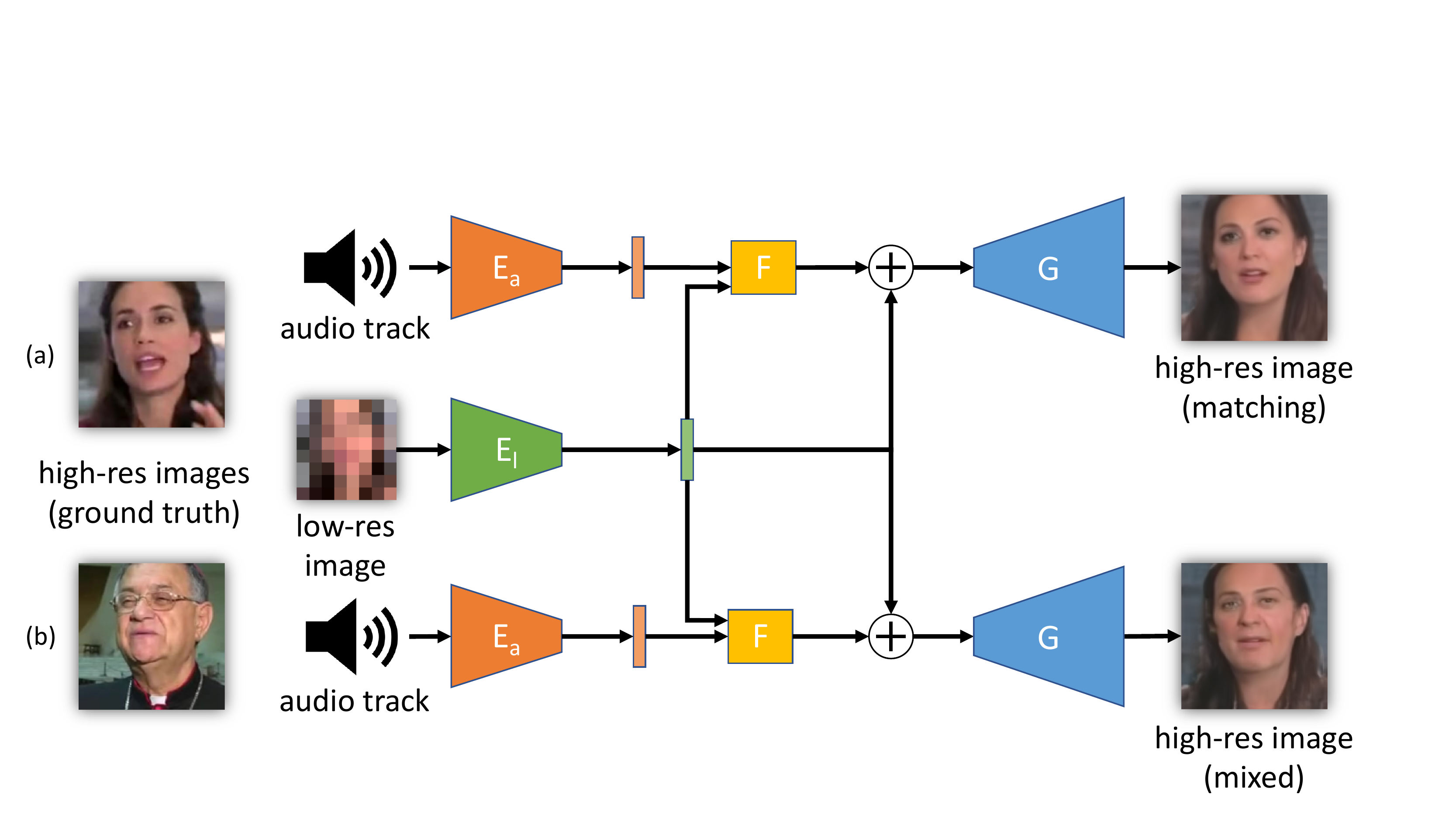}
    \caption{Simplified training and operating scheme of the proposed model. The model can be used (a) with matching inputs or (b) by mixing low-resolution images with audios from other videos. The low-resolution image ($8\times 8$ pixels) is fed to a low-resolution encoder $E_l$ to obtain an intermediate latent representation. A residual is computed by fusing in the network $F$ the encoded audio track (through the audio encoder $E_a$) with the encoded low-resolution image. The residual is used to update the latent representation of the low-resolution image and then produce the high-resolution image through the generator $G$. The images to the right are actual outputs of our trained model.}
  	\label{fig:model}
\end{figure*}

Therefore, we propose to build a model for face super-resolution by exploiting both a low-resolution image and its audio.
To the best of our knowledge, this has never been explored before.
A natural way to solve this task is to build a \emph{multimodal network} with two encoding networks, one for the low-resolution image and one for audio, and a decoding network mapping the concatenation of the encoders outputs to a high resolution image. 
In theory, a multi-modal network should outperform its uni-modal counterparts. In practice, however, this does not happen with standard networks and training strategies, as shown empirically in \cite{DBLP:journals/corr/abs-1905-12681}. 
According to \cite{DBLP:journals/corr/abs-1905-12681} the performance gap is due to: 1) the difference between modalities in term of convergence and over-fitting speeds, 2) The susceptibility of multi-modal architectures to over-fitting due to their higher capacity. 
To address the above training issues of multi-modal networks, we propose to train the low-resolution image encoder and the audio encoder separately, so that their disentanglement accuracy can be equalized. 
To this aim, we first train a generator 
that starts from a Gaussian latent space and outputs high resolution images (see Fig.~\ref{fig:model}). The generator is trained as in the recent StyleGAN of \cite{karras2018style}, which produces very high quality samples and a latent space with a useful hierarchical structure. Then, we train a reference encoder to invert the generator by using an autoencoding constraint. The reference encoder maps a high-resolution image to the latent space of the generator, which then outputs an approximation of the input image. Then, given a matching high/low-resolution image pair, we pre-train a low-resolution image encoder 
to map its input to the same latent representation of the reference encoder (on the high-resolution image). As a second step, we train an audio encoder 
and a fusion network to improve the latent representation of the (fixed) low-resolution image encoder.
To speed up the training of the audio encoder we also pre-train it by using as latent representation the average of the outputs of the reference encoder on a high-resolution image and its horizontally mirrored version. Thanks to the hierarchical structure of the latent space learned through StyleGAN, this averaging removes information, such as the viewpoint, that  audio cannot possibly carry. In Section~\ref{sec:method}, we describe in detail the training of each of the above models. Finally, in Section~\ref{sec:exp} we demonstrate experimentally that the proposed architecture and training procedure successfully fuses aural and visual data. We show that the fusion yields high resolution images with more accurate identities, gender and age attributes than the reconstruction based on the lone low-resolution image. We also show that the fusion is semantically meaningful by mixing low-resolution images and audio from different videos 
(see an example in Fig.~\ref{fig:model}~(b)).

\noindent\textbf{Contributions:}
Our method builds three models for the following mappings: 1) Audio to high-resolution image; 2) Low-resolution image to high-resolution image; 3) Audio and low-resolution image to high-resolution image. The first mapping was developed concurrently to Speech2Face \cite{oh2019speech2face}. A notable difference is that Speech2Face is trained using as additional supervision a pre-trained face recognition network, while our method is fully unsupervised. In the second mapping, we show in our Experiments section that we achieve state of the art performance at $16\times$. In the last mapping, which is the main novelty of this paper, we show that our trained model is able to transfer and combine facial attributes from audio and low-resolution images.\footnote{Our code and pre-trained models are available at \href{https://gmeishvili.github.io/ear_for_face_super_resolution/index.html}{https://gmeishvili.github.io/ear\_for\_face\_super\_resolution/index.html}.}


\section{Related Work}
\noindent\textbf{General Super-Resolution. }
Singe Image Super Resolution (SISR) is a very active research area, which largely benefitted from the latest developments in deep learning (see, \eg, \cite{He_2019_CVPR,Haris_2019_CVPR,Zhang_2019_CVPR_1,Deng_2019_ICCV,Kim_2019_ICCV}).
A wide set of instances of this problem has been addressed, ranging from arbitrary scale factors \cite{Hu_2019_CVPR}, to improving the realism of the training set through accurate modeling \cite{Xu_2019_CVPR,Cai_2019_ICCV} or through using real zoomed-in images \cite{Chen_2019_CVPR,Zhang_2019_CVPR}, to robustness against adversarial attacks \cite{Choi_2019_ICCV} and generalization \cite{Zhou_2019_ICCV}, and to modeling multiple degradations \cite{Zhang_2018_CVPRR,Gu_2019_CVPR,Zhang_2019_CVPR}. Finally, 
\cite{Rad_2019_ICCV,Soh_2019_CVPR} focus on the evaluation of the image quality in SISR.
Advances in general super-resolution have also been largely driven by the introduction of task-specific network architectures and components (see \eg, \cite{Zhang_2018_CVPR, Li_2018_ECCV, Ahn_2018_ECCV, Tai_2017_CVPR, Zhang_2018_ECCV, Hui_2018_CVPR, Wang_2018_CVPR, Haris_2018_CVPR, Han_2018_CVPR, Lai_2017_CVPR, Li_2019_CVPR_1, Wang_2019_CVPR, Dai_2019_CVPR, Yi_2019_ICCV, Qiu_2019_ICCV, Zhang_2019_ICCV,Li_2019_CVPR}).
In our method, we do not rely on task-specific architectures, although we leverage the design of a state-of-the-art generative model \cite{karras2018style}. 

\noindent\textbf{Face Super-Resolution. }
The face super-resolution problem has been tackled with a wide variety of approaches. For example, Huang \etal~\cite{Huang_2017_ICCV} trained a CNN to regress wavelet coefficients of HR faces and Yu \etal~\cite{Yu_2017_CVPR} introduced a transformative discriminative autoencoder to super-resolve unaligned and noisy LR face images. 
More in general, recent methods addressed the problem by using additional supervision, for example, in the form of facial landmarks, heat-maps or the identity label, and multi-task learning (\eg, \cite{Bulat_2018_CVPR,Yu_2018_ECCV,Chen_2018_CVPR,Zhang_2018_ECCV_1,Yu_2018_CVPR}). 
In contrast, by using videos with corresponding audio tracks our method does not rely on additional human annotation and thus its training can scale more easily to large datasets.\\
\noindent\textbf{GAN-Based Super-Resolution. }
Many general super-resolution methods also make use of adversarial training (see, \eg, \cite{Ledig_2017_CVPR,Park_2018_ECCV,Bulat_2018_ECCV,Zhang_2019_ICCV_1}).
Several super-resolution methods based on Generative Adversarial Networks (GANs) \cite{NIPS2014_Goodfellow} focus specifically on faces \cite{Bulat_2018_CVPR,Chen_2018_CVPR,Yu_2018_CVPR,Xu_2017_ICCV}. 
Our work also relies on the use of a GAN to learn a face specific prior. 
However, our approach builds a more general generative network that combines low-resolution images and audio (see Fig.~\ref{fig:model}).\\
\noindent\textbf{Use of Audio in Vision Tasks. }
The use of audio in combination with video has received a lot of attention recently (see, \eg, \cite{song2018talking,zhu2018high}).
Audio and video have been combined to learn to localize objects or events \cite{Arandjelovic_2018_ECCV,Tian_2018_ECCV}, to learn how to separate audio sources \cite{Owens_2018_ECCV,Zhao_2018_ECCV,Gao_2018_ECCV,ephrat2018looking}, to learn the association between sound and object geometry and materials \cite{Sterling_2018_ECCV}, and to predict body dynamics \cite{Shlizerman_2018_CVPR}.
A significant body of work has also been devoted to the mapping of audio to visual information (see, \eg, \cite{oh2019speech2face} and references therein).
However, to the best of our knowledge we are the first to combine audio and images for an image restoration task.

\section{Extreme Face Super-Resolution with Audio}
\label{sec:method}
Our goal is to design a model that is able to generate high resolution images based on a (very) low resolution input image and an additional audio signal. 
The dataset is therefore given by $\mathcal{D} = \big\{ (x_i^{h}, x_i^{l}, a_i)\mid i=1, \ldots, n \big\}$ where $x_i^{h}$ is the high-resolution image, $x_i^{l}$ is the low-resolution image and $a_i$ is a corresponding audio signal. 
Our model consists of several components: a low-resolution encoder $E_l$, an audio encoder $E_a$, a fusion network $F$ and a face generator $G$. 
An overview of the complete architecture is given in Fig.~\ref{fig:model}.
\subsection{Combining Aural and Visual Signals}
As mentioned in the Introduction, a natural choice to solve our task is to train a feedforward network to match the ground truth high resolution image given its low-resolution image and audio signal.
Experimentally, we found that such a system tends to ignore the audio signal and to yield a one-to-one mapping from a low-resolution to a single high-resolution image. 
We believe that this problem is due to the different nature of the aural and visual signals, and the choice of the structure of the latent space.
The fusion of both signals requires mapping their information to a common latent space through the encoders.
However, we find experimentally that the audio signal requires longer processing and more network capacity to fit the latent space (this is also observed in \cite{DBLP:journals/corr/abs-1905-12681}).
This fitting can also be aggravated by the structure of the latent space, which might be biased more towards images than audio.
Ideally, the low-resolution image should only condition the feedforward network to produce the most likely corresponding high-resolution output and the audio signal should 
introduce some local variation (\ie , modifying the gender or the age of the output). 
Therefore, for the fusion to be effective it would be useful if the audio could act on some fixed intermediate representation from the low-resolution image, where face attributes present in the audio are disentangled.

For these reasons we opted to pre-train and fix the generator of a StyleGAN \cite{karras2018style}
and then to train encoders to autoencode the inputs by using the generator as a decoder network.
StyleGAN generators have been shown to produce realistic high resolution images along with a good 
disentanglement of some meaningful factors of variation in the intermediate representations. 
Such models should therefore act as good priors for generating high resolution face images and 
the disentangled intermediate representations should allow better editing based on the audio signal. 
Formally, we learn a generative model of face images $G(z)$, where $z \sim \mathcal{N}\left(0, I_{d}\right)$, 
by optimizing the default non-saturating loss of StyleGAN (see \cite{karras2018style} for details).
\subsection{Inverting the Generator}

Our goal is that the fusion of the information provided by the low-resolution image and audio track results in a reconstruction that is close to the corresponding high resolution image. We pose this task as that of mapping an image $x$ to its latent space target $z$, such that $G(z) = x$. In other words, we need to invert the pre-trained generator $G$.
Recently, this problem has attracted the attention of the research community \cite{Bau_2019_ICCV}. 
In this paper, we introduce a novel GAN inversion approach, where we first pre-train the encoder $E_h$ while the generator is fixed, and then we train both the encoder $E_h$ and the generator $G$ (fine-tuning) through an autoencoding constraint and by anchoring the weights of $G$ to its initial values through an $L_2$ loss. Then, the latent representation $z_i$ corresponding to the image $x_i$ can be generated by the encoder $E_h$, and used as a target by the encoders of the low-resolution images and the audio, and the fusion network.

\noindent\textbf{Encoder Pre-Training.} As a first step we train a high-resolution image encoder by minimizing the loss
\begin{align}
	{\cal L}_\text{pre-train} = \sum_{i=1}^n \left|G( z_i)-x_i^h\right|_1 + \lambda_{f}\ell_\text{feat}\left(G( z_i ), x_i^h\right),
		\label{eq:fixed_generator}
\end{align}
only with respect to $E_{h}$, where $z_i = E_{h}(x_i^h)$, $\ell_\text{feat}$ is a perceptual loss based on VGG features (see Supplementary material for more details), and $\lambda_f=1$ is a coefficient that regulates the importance of $\ell_\text{feat}$ relative to the $L_1$ loss.
We found that regressing a single $z_i$ is not sufficient to recover a good approximation of $x_i^h$. 
In the original style-based generator \cite{karras2018style} each $z_i$ is mapped to a vector $w_i$, which is then replicated and inserted at $k$ different layers of the generator (each corresponding to different image scales). 
To improve the high-resolution reconstruction, we instead generate $k$ different $z_{ij}$, $j=1,\dots,k$, and feed the resulting $w_{ij}$ to the corresponding layers in the generator. The output of $E_h$ therefore lies in $\mathbb{R}^{k \times d}$. 
Note that this is not too dissimilar from the training of the style-based generator, where the $w$-s of different images are randomly mixed at different scales.
\begin{figure}[t]
    \centering
    \includegraphics[width=0.99\linewidth]{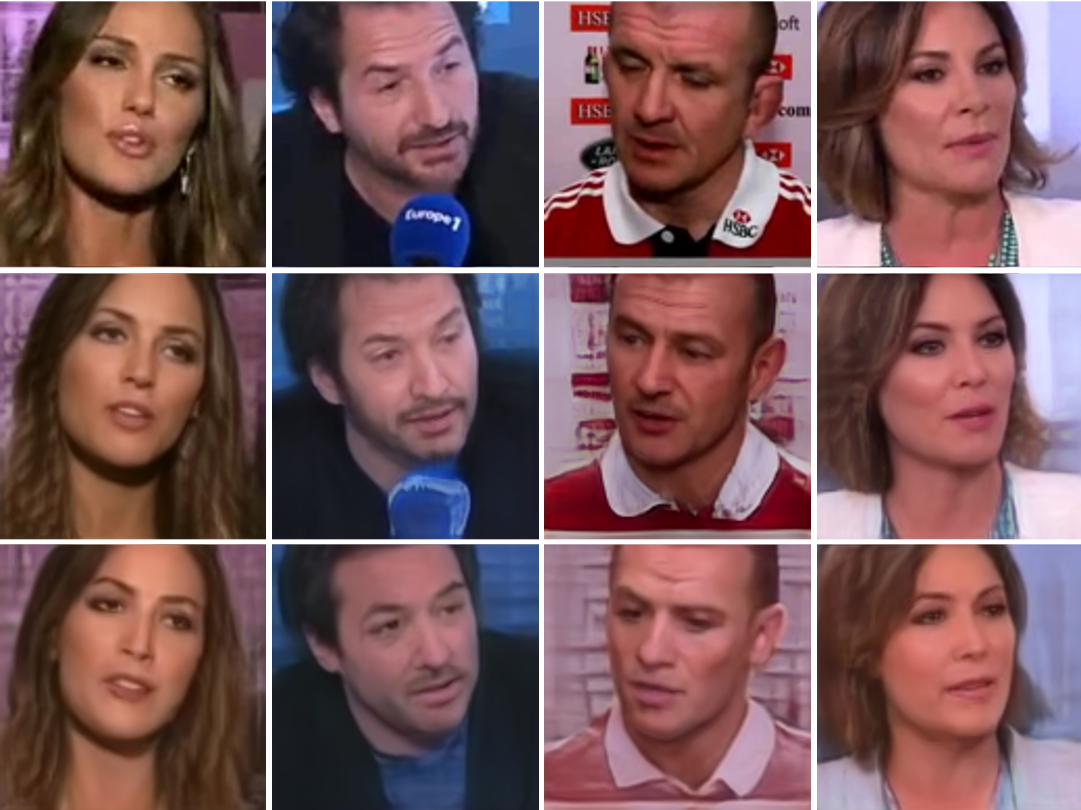}     
    \caption{Examples of generator inversions. Top row:  Input images to the autoencoders. Bottom row: Autoencoding results with a fixed pre-trained generator (see eq.~\eqref{eq:fixed_generator}). Middle row: Autoencoding results with our fine-tuned generator (see eq.~\eqref{eq:fine_tuned_generator}). }
  	\label{fig:inversion}
\end{figure}

\noindent\textbf{Encoder and Generator Fine-Tuning.}
This second optimization problem can be written as
\begin{align}
	\small
	\min _{E_{h},G} & {\cal L}_\text{pre-train}
	+\lambda_{t}\left|G_\text{init} - G\right|_2^2,
  	\label{eq:fine_tuned_generator}
\end{align}
where $\lambda_t=1$ is a coefficient that regulates how much $G$ can be updated, and $G_\text{init}$ denotes the weights of $G$ after StyleGAN training.
Moreover, during training we relax the regularizer of the weights of $G$ by reducing $\lambda_{t}$ by a factor of 2 as soon as the overall loss is minimized (locally). The purpose of the pre-training and the regularizer decay procedure is to encourage a gradual convergence of both the encoder and the decoder without losing prematurely the structure of the latent representation of $G$. 
Examples of inversions before and after the fine-tuning are shown in Fig.~\ref{fig:inversion}. There is a visible improvement in the reconstruction accuracy of both the face and the background. Quantitative results are shown in the Experiments section.
\begin{figure}[t]
    \centering
    \includegraphics[width=1\linewidth,trim=0.9cm 0 1.1cm 0, clip]{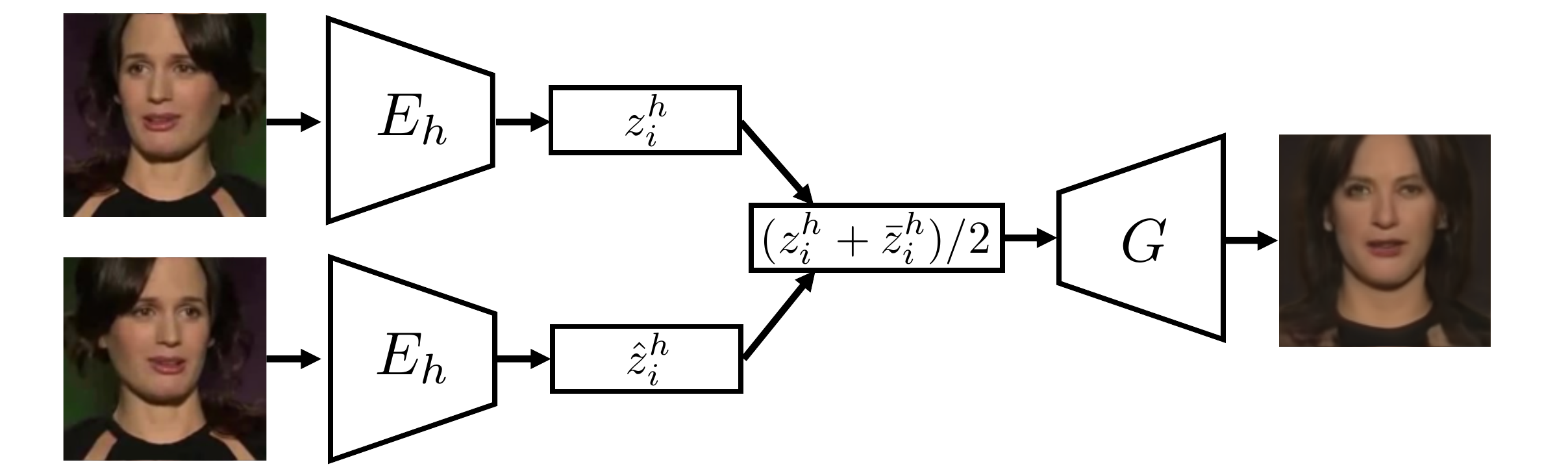}     
    \caption{Illustration of how we compute the targets for the audio encoder pre-training. We feed a high-resolution training image and its horizontally flipped version through the high-resolution encoder. The resulting latent codes are then averaged and used as targets. Because of the hierarchical structure of the latent space of StyleGAN, the averaged latent code produces a face in neutral frontal facing pose.}
  	\label{fig:ea}
\end{figure}
\subsection{Pre-Training Low-Res and Audio Encoders}
Given the high-resolution image encoder, we now have targets $z_i$ for the low-resolution and audio fusion. 
However, the training of a fusion model directly on these targets runs into some difficulties.
As mentioned before, we find experimentally that, given enough capacity, a fusion model $F(x_i^l, a_i)$ trained to predict $z_i = E_{h}(x_i^h)$, ignores the audio signal $a_i$ almost completely. 
To address this degenerate behavior, we train two encoders $E_l$ and $E_a$ separately to extract as much information from the two modalities as possible and only later fuse them.
To ensure that neither of the two encoders can overfit the whole training set $\mathcal{D}$ we extract the subset $\mathcal{D}_\text{pre} = \big\{ (x_i^{h}, x_i^{l}, a_i)\mid i=1, \ldots, n/2 \big\}$ for the encoders pre-training and use the entire $\mathcal{D}$ only for the later fusion training.
The low-resolution encoder $E_l$ is trained to regress the high-resolution encodings $z_i=E_{h}(x_i^h)$ from $x_i^l$ by solving
\begin{align}
	\small
	\min _{E_{l}}\!\!\!\!\sum_{x_i^l, x_i^h \in \mathcal{D}_\text{pre}}\!\!\!\!\!\!  \left|E_{l}\left(x_i^l\right) - z_i\right|_1+ \lambda \left|D\circ G\left( E_{l}\left(x_i^l\right)\right) - x_i^l\right|_1,
\end{align}
where $D\circ x$ is the $16\times$ downsampling of $x$ and $\lambda=40$.

In the case of the audio encoding, regressing all the information in $z_i$ with $E_a(a_i)$ is not possible without overfitting, as many of the factors of variation in $z_i$, \eg, the pose of the face, are not present in $a_i$. 
To remove the pose from $z_i$ we generate the targets for the audio encoder as $\bar{z_i}\doteq\frac{1}{2}\big( E_h(x_i^h)+E_h(\hat{x}_i^h)\big)$, where $\hat{x}_i^h$ is a horizontally flipped version of the image $x_i^h$. As it turns out, due to the disentangled representations of $G$, the reconstruction $G(\bar{z_i})$ produces a neutral frontal facing version of $G(z_i)$ (see Fig.~\ref{fig:ea}). The audio encoder $E_a$ is finally trained by solving
\begin{equation}
	\min _{E_{a}} \sum_{a_i, x_i^h \in \mathcal{D}_\text{pre}}   \left|E_{a}(a_i) - \bar{z_i}\right|_1 .
\end{equation}
\subsection{Fusing Audio and Low-Resolution Encodings}
We now want to fuse the  information provided by the pre-trained encoders $E_l$ and $E_a$. 
Since the low-resolution encoder $E_l$ already provides a good approximation to $E_h$, it is reasonable to use it as a starting point for the final prediction. 
Conceptually, we can think of $E_l$ as providing a $z_i^l=E_l(x_i^l)$ that results in a canonical face $G(z_i^l)$ corresponding to the low-resolution image $x_i^l$.
Ambiguities in $z_i^l$ could then possibly be resolved via the use of audio, which would provide an estimate of the residual $\Delta z_i = z_i-z_i^l$.
We therefore model the fusion mechanism as $z_i^f=E_l(x_i^l)+F\big(E_l(x_i^l), E_a(a_i)\big)$, where $F$ is a simple fully-connected network acting on the concatenation of $E_l(x_i^l)$ and $E_a(a_i)$. 
Since the audio-encoding $E_a$ might be suboptimal for the fusion, we continue training it along with $F$. 
The limited complexity of the function $F$ prevents the overfitting to the low-resolution encoding, but provides the necessary context for the computation of $\Delta z_i$. To summarize, we train the fusion by optimizing 
\begin{align}
	\min _{E_{a}, F}\!\!\!\!\sum_{a_i, x_i^h, x_i^l \in \mathcal{D}}\!\! \left|z_i^f - z_i\right|_1
	+ \lambda \left|D\circ  G\left( z_i^f\right) - x_i^l\right|_1.
\end{align}
\vspace{-1em}
\begin{figure}[t]
\centering
\includegraphics[width=1.0\linewidth,trim=0.4cm 0 0.3cm 0,clip]{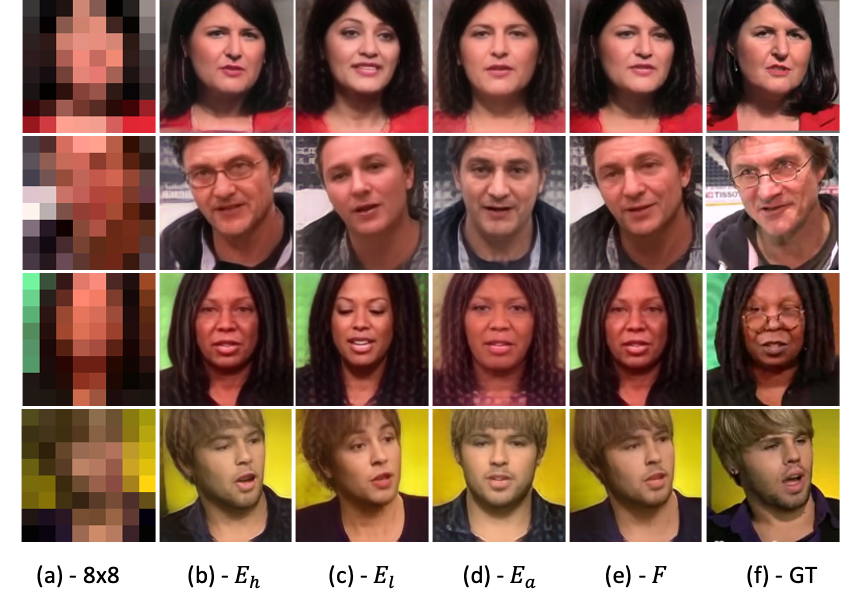}
\caption{Selected examples of reconstructions to some of our ablation experiments. The $8\times8$ pixels low-resolution inputs are shown in (a) and the corresponding $128\times128$ pixels ground truth images are shown in column (f). In the middle, we show results for encodings from the high-resolution encoder $E_h$ in (b), the low-resolution encoder $E_l$ in (c), the audio encoder $E_a$ in (d) and from our fusion model $F$ with the fine-tuned $E_a$ in (e). It is apparent how our integration of the audio and the low-resolution image is limited by the accuracy of the encoder for the high resolution image (compare \eg, (b) with (e) and (f) on the third row).
	\label{fig:abl}
}
\end{figure}
\begin{figure}[t]
    \centering
    \includegraphics[width=0.99\linewidth]{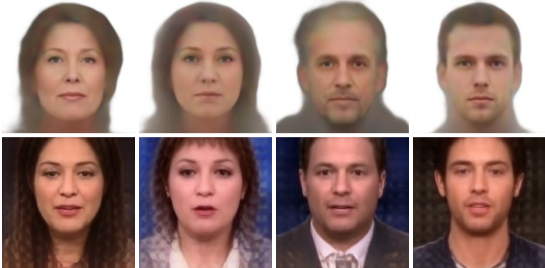}     
    \caption{To demonstrate qualitatively the capabilities of our audio-to-image model $E_{a}+G$ we picked several audio tracks and the corresponding generated faces by Oh \etal~\cite{oh2019speech2face} from \url{https://speech2face.github.io/supplemental/retrieve/index.html}. Images in every column are generated from the same audio sources. The results of Oh \etal~\cite{oh2019speech2face} are shown on the first row, and our results are shown on the second row.}
  	\label{fig:speech2face}
\end{figure}

\section{Experiments}\label{sec:exp}

We demonstrate our contributions by evaluating three models with different input-output mappings: 
1) Audio to high-resolution image; 
2) Low-resolution image to high-resolution image; 
3) Audio and low-resolution image to high-resolution image.
In particular, we focus our attention on the third case as it is the main objective of this paper.

\subsection{Dataset}
We perform all our experiments on a subset of the VoxCeleb2 dataset \cite{Chung18b}. 
The dataset contains over one million audio tracks extracted from 145K videos of people speaking. 
For the full training set $\mathcal{D}$ we select 104K videos with 545K audio tracks and extract around 2M frames at $128 \times 128$ pixels such that each speaker has at least 500 associated frames. 
We then extract half of this dataset to create $\mathcal{D}_\text{pre}$ in such a way that $\mathcal{D}_\text{pre}$ and $\mathcal{D}$ contain the same speakers, but $\mathcal{D}_\text{pre}$ has fewer videos than $\mathcal{D}$. 
For the test set, we select 39K frames and 37K utterances from 25K videos not contained in the training set (again from the same speakers).
In the end, we select around 4K speakers out of the 6K speakers in the full dataset (filtering out speakers with very few videos and audio tracks).
Note that this selection is purely done to allow the evaluation via a speaker identity classifier.  
We call experiments \textbf{closed set} when the training and test sets share the same set of face identities; instead, we call them \textbf{open set} when the test set has identities that were not in the training set.
\subsection{Implementation}
The style-based generator $G$ is pre-trained on the full training set $\mathcal{D}$ with all hyper-parameters set to their default values (see \cite{karras2018style} for details). 
It has been trained on a total of 31 million images. 
The high-resolution encoder $E_h$ is trained for 715K iterations and a batch-size of 128 on the $128 \times 128$ images from $\mathcal{D}$.
The low-resolution encoder $E_l$ and the audio encoder $E_a$ are trained on $\mathcal{D}_\text{pre}$. 
$E_l$ is trained for 240K iterations with a batch-size of 256 and $E_a$ is trained for 200K iterations and a batch-size of 64. 
The inputs $x_i^l$ to $E_l$ are of size $8 \times 8$ pixels and the inputs to $E_a$ are the audio log-spectrograms of $a_i$ of size $257 \times 257$.  
The fine-tuning of $E_a$ and the training of the fusion layer $F$ is performed for 420K iterations on  $\mathcal{D}$.
We use the Adam optimizer \cite{kingma2014adam} with a fixed learning rate of $10^{-4}$ for the training of all the networks. 
A detailed description of the network architectures can be found in the supplementary material. During the training of $F$ we sample one frame and an audio segment (4 seconds) independently (\ie, they are not synchronized) from the same short video. This forces the network to learn more general attributes such as gender and age, rather than characteristics specific to a particular instant in time.

\subsection{Audio-Only to High Resolution Face}

Although our main objective is to obtain super-resolved images from the fusion of low-resolution images and audio, we provide a brief comparison between our model for face reconstruction from audio ($E_a+G$) with Speech2Face \cite{oh2019speech2face}. Since the dataset of \cite{oh2019speech2face} is not public, we perform a qualitative and a quantitative comparison based on audio tracks and reconstrucitons by Oh \etal~\cite{oh2019speech2face} from \url{https://speech2face.github.io/supplemental/retrieve/index.html}. In Fig.~\ref{fig:speech2face} we show the reference faces obtained by Speech2Face and our output using the same audio tracks. We can see that the gender and age are matching. In the second evaluation, we perform gender classification on the output of our audio-to-image model when given audio from the VoxCeleb dataset \cite{Chung18b} as input. Given a voice of the male or female person, our $E_{a}+G$ model generates faces of males and females in $97\%$ and $96\%$ of the cases respectively. The results match those reported by \cite{oh2019speech2face}. Notice that \cite{oh2019speech2face} uses supervision from a classifier during training, while our training is completely unsupervised. 
 
 \begin{table}
\caption{Results of our ablation experiments. We report the accuracy of an identity classifier $C_i$ and a gender classifier $C_g$ as well as the error of an age classifier $C_a$ on generated high-resolution images. 
All the models in (c)-(h) are trained using the fine-tuned generator $G$. \label{tab:abl}}
\centering
\footnotesize
\begingroup
\setlength{\tabcolsep}{2.45pt}
\begin{tabular}{@{}lcc@{\hspace{1mm}}c@{\hspace{1mm}}cc@{\hspace{1mm}}c@{}}\toprule
 & \multicolumn{3}{c}{\textbf{Closed Set}} & \multicolumn{3}{c}{\textbf{Open Set}}\\ 
\textbf{Ablation} 		& \textbf{Acc $C_i$} & \textbf{Acc $C_g$} & \textbf{Err $C_a$} & \textbf{Acc $C_i$} & \textbf{Acc $C_g$} & \textbf{Err $C_a$} \\ \midrule
(a) $E_{h}$ + fixed $G$ 			& 34.31\% & 95.60\%  & 3.59 & 29.42\% & 92.65\% & 3.28\\
(b) $E_{h}$ + tuned $G$ 			& 71.62\% & 98.20\% & 2.85 & 64.95\% & 95.14\% & 2.74\\ \midrule
(c) $E_{l}$ only 			& 36.47\% & 95.51\% & 3.62 & 15.55\% & 91.08\% & 3.76\\
(d) $E_{a}$ only 			& 26.06\% & 97.07\% & 4.29 & \phantom{0}0.20\% & \textbf{96.38}\% & 4.85\\ 
(e) $F_1$ + tuned $E_a$ 					& 35.91\% & 95.88\% &	3.56 & 15.03\% & 91.75\% & \textbf{3.64}\\  
(f) $F$ + zero $E_a$ 		& 36.95\% & 95.53\% &	3.60 & 15.38\% & 90.89\% & 3.73\\
(g) $F$ + fixed $E_a$ 		& 48.43\% & 97.17\% & 3.46 & 14.57\% & 92.86\% & 3.74\\
(h) $F$ + tuned $E_a$				& \textbf{51.65\%} & \textbf{97.32\%}	& \textbf{3.31} & \textbf{15.67\%} & 93.11\% & 3.68\\\bottomrule
\end{tabular}
\endgroup
\end{table}
 
\subsection{
Classification 
as a Performance Measure}
\begin{figure}[]
    \centering
    \includegraphics[width=1\linewidth,trim=.2cm 0 0 0, clip]{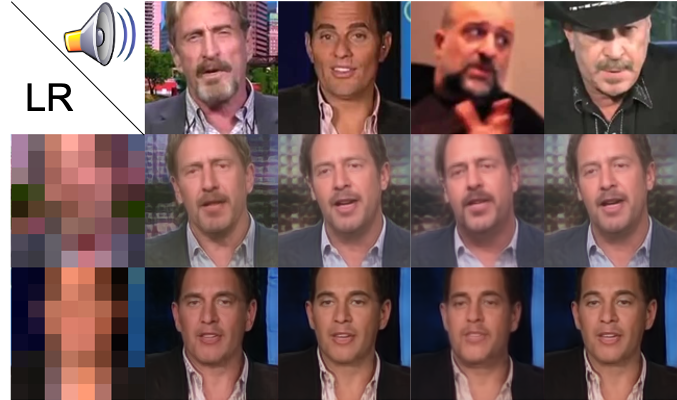}     
    \caption{
    Examples where we mix a given low-resolution image with different audio sources. The top row shows the high-resolution images from which we take the audio track. The first column on the left corresponds to the low-resolution images used as input. The rest of the images in the matrix are generated by mixing the low-resolution image from a row with the audio of a column.}
  	\label{fig:mix}
\end{figure}

To evaluate the capability of our model to recover gender and other identity attributes based on the low-resolution and audio inputs we propose to use the accuracy of a pre-trained identity classifier $C_i$ and gender classifier $C_g$ which achieve an accuracy of $95.25\%$ and $99.53\%$ respectively on the original high resolution images.
To this end, we fine-tune two VGG-Face CNNs of \cite{Parkhi15} on the training set $\mathcal{D}$ for 10 epochs on both face attributes. 
As one can see in Table~\ref{tab:abl} these classifiers perform well on the test set on both face attributes. Although we do not have the ground truth age of our dataset, we use a pre-trained age classifier $C_a$ \cite{Rothe2018} as the reference. Then, we measure the performance of our models by checking the consistency between the classified age of the input and of the output.\\
\begin{table*}
\caption{Comparison to other general-purpose super-resolution methods at different super-resolution factors. We report PSNR and SSIM obtained on the test set. Note that the target resolution is fixed at $128\times128$ pixels and therefore the inputs to the $4\times$ method (top row, LapSRN \cite{Lai_2017_CVPR}) is $32 \times 32$ pixels, while our model only uses $8\times8$ pixels input images.}\label{tab:comp}
\centering
\begingroup
\setlength{\tabcolsep}{2.45pt}
\begin{tabular}{@{}lc@{\hspace{3em}}ccccc@{\hspace{3em}}ccccc@{}}
\toprule
& & \multicolumn{5}{c}{\textbf{Closed Set}} & \multicolumn{5}{c}{\textbf{Open Set}}\\ 
\textbf{Method}&\textbf{Factor} & \textbf{PSNR} & \textbf{SSIM} & \textbf{Acc $C_i$} & \textbf{Acc $C_g$} & \textbf{Err $C_a$} & \textbf{PSNR} & \textbf{SSIM} & \textbf{Acc $C_i$} & \textbf{Acc $C_g$} & \textbf{Err $C_a$} \\ \midrule
LapSRN  \cite{Lai_2017_CVPR} 	& $4\times$ & 31.99 & 0.91 & 93.83\% & 99.38\% & 2.81 & 31.66 & 0.91 & 95.84\% & 95.37\% & 2.81 \\ \midrule
LapSRN \cite{Lai_2017_CVPR} 	& $16\times$ & 22.75 & 0.64 	 & \phantom{0}5.27\% & 83.27\% & 5.16 & 22.39 & 0.62 & \phantom{0}6.80\% & 79.57\%	& 5.16\\
W-SRNet \cite{Huang_2017_ICCV} & $16\times$ & 21.55 & 0.67 & 34.91\% & 95.68\% & 4.28 & 19.18 & 0.59 & 13.54\% & 89.45\% & 4.57 \\ 
\textbf{Ours}	& $16\times$ & 21.64  &	0.68 	& 51.65\% & 97.32\%	& 3.31 & 19.97 & 0.60 & 15.67\% & 93.11\% & 3.68\\ \bottomrule
\end{tabular}
\endgroup
\end{table*}
\noindent\hspace{-0.4em}\textbf{Ablations.} We perform ablation experiments to understand the information retained in the encoders and to justify the design of our final model. The accuracy of the classifiers $C_i$ and $C_g$, as well as the consistency error of $C_a$, are reported in Table~\ref{tab:abl} for the following ablation experiments
\begin{description}
\item[(a)-(b) The importance of fine-tuning.] In (a) we show the performance after pre-training of $E_h$ without fine-tuning, and in (b) we show the improvement in performance with the fine-tuning of $G$ as in eq.~\eqref{eq:fine_tuned_generator}.
\item[(c)-(d) Individual components.] Shows the performance of the individual encoders without fusion. Results for the low-resolution encoder $E_l$ and the audio encoder $E_a$ should be compared to the reference high-resolution encoder $E_h$. 
\item[(e)-(h) Fusion strategies.] The performance of different fusion strategies is reported. As a reference, we report results of a fusion model $F_1$ with a single fully-connected layer and fine-tuning of $E_a$. We compare it to a more complex fusion network $F$ with three fully-connected layers when the audio is not used (f), the audio encoder is fixed (g) and when fine-tuning $E_a$ (h).
\end{description}
Ablation (c) and (d) show that $E_a$ is able to predict the correct gender more often than $E_l$. 
All the fusion approaches (e)-(h) lead to an improvement in terms of identity prediction over $E_a$ and $E_l$ alone, thus showing that the information from both inputs is successfully integrated. We can observe that both gender and age can be predicted well even from unseen identities (\ie, the open-set case). Ablation (f) vs (h) shows that the method is indeed using information from the audio signal and the performance increase is not due to the additional capacity of the fusion network $F$. Ablation (h) vs (e) justifies the usage of 3 fully-connected layers instead of only 1. Ablation (h) vs (g) demonstrates that fine-tuning of the encoder $E_{a}$ during the training of the fusion network $F$ leads to slight improvements in terms of our quantitative metrics. Note that the performance of all methods in Table~\ref{tab:abl}, including the SotA \cite{Huang_2017_ICCV}, is lower in the open set experiments than in the closed set ones. This is expected since all methods are trained only on identities present in the training set and most likely only a small amount of information is shared across identities. The open set experiments show how much the methods can identify such shared information, which is a sign of generalization.
See also Fig.~\ref{fig:abl} for qualitative results.\\
\begin{figure}[]
    \centering
    \includegraphics[width=1.0\linewidth]{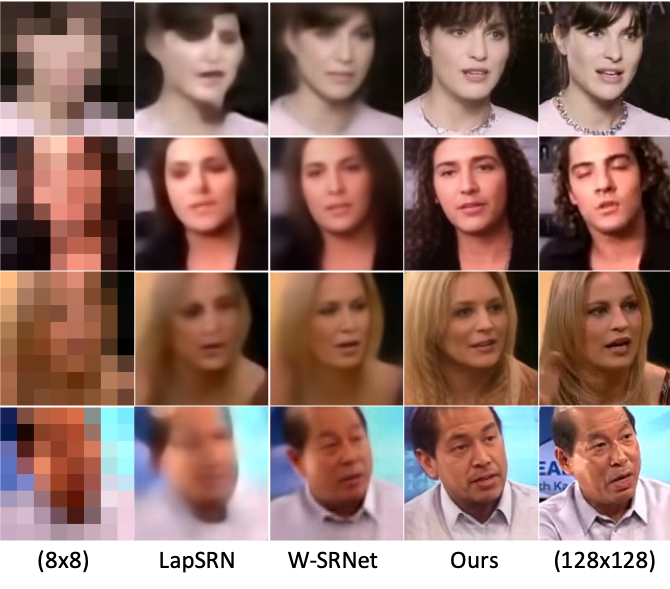}
    \caption{Comparison to other super-resolution methods on our test set. The first column shows the $8\times8$ pixels inputs; the second column shows the output of LapSRN \cite{Lai_2017_CVPR}; the third column shows the output of W-SRNet \cite{Huang_2017_ICCV}. Our model is shown in the fourth column. The ground-truth high-resolution image is shown in the last column.
    }
  	\label{fig:comp}
  	\vspace{-4mm}
\end{figure}
\noindent\hspace{-0.6em}\textbf{Comparisons to Other Super-Resolution Methods.}
We compare to state-of-the-art super-resolution methods in Table~\ref{tab:comp} and Fig.~\ref{fig:comp}. 
The standard metrics PSNR and SSIM along with the accuracy of $C_i$ and $C_g$, and the errors of $C_a$ are reported for super-resolved images of our test set. 
Note that most methods in the literature are not trained on extreme super-resolution factors of $16\times$, but rather on factors of $4\times$. Therefore, we report the results of one method using a factor of $4\times$ as a reference for the changes with the $16\times$ factor. The performance of other $4\times$ super-resolution methods can be found in our supplementary material.
We retrain the methods of \cite{Lai_2017_CVPR} and \cite{Huang_2017_ICCV} on our training set, before evaluating their performance. Notice that although LapSRN trained on $16\times$ super-resolution performs better in terms of PSNR and SSIM than our method, the quality of the recovered image is clearly worse (see Fig.~\ref{fig:comp}). This difference in the quality is instead revealed by evaluating the gender and identity classification accuracies, and the age classification error of the restored images. This suggests that while PSNR and SSIM may be suitable metrics to evaluate reconstructions with small super-resolution factors, they may not be suitable to assess the reconstructions in more extreme cases such as with a factor of $16\times$.
\begin{table}[t]
\caption{Agreement of $C_g$ predictions with labels of low-resolution and audio labels on mixed reconstructions. 
\label{tab:audio_correlation}}
\centering
\begingroup
\setlength{\tabcolsep}{2.45pt}
\begin{tabular}{@{}lcc@{\hspace{1mm}}c@{\hspace{1mm}}cc@{\hspace{1mm}}c@{}}\toprule
\textbf{Label Source} & \textbf{Closed Set} & \textbf{Open Set}\\ \midrule
 Audio  			& 10.76\% & 13.74\%\\
Low-Resolution Image & 89.24\% & 86.26\% \\\bottomrule
\end{tabular}
\endgroup
\end{table}
\begin{figure}[]
    \centering
    \includegraphics[width=1.0\linewidth]{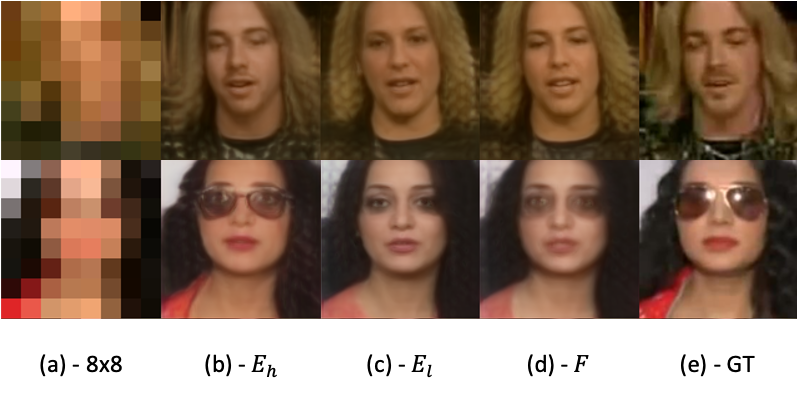}     
    \caption{
    Examples of failure cases in our method. The $8\times8$ pixels low-resolution inputs are shown in (a) and the corresponding $128\times128$ pixels ground truth images are shown in column (e). In the middle, we show results for encodings from the high-resolution encoder $E_h$ in (b), the low-resolution encoder $E_l$ in (c) and from our fusion model $F$ with the fine-tuned $E_a$ in (d).}
  	\label{fig:failure}
\end{figure}\\
\noindent\textbf{Editing by Mixing Audio Sources. }
Our model allows us to influence the high-resolution output by interchanging the audio tracks used in the fusion.
To demonstrate this capability we show examples where we mix a fixed low-resolution input with several different audio sources in Fig.~\ref{fig:mix}. 
To also quantitatively evaluate such mixing, we feed low-resolution images and audios from persons of different gender and classify the gender of the resulting high-resolution faces. In Table~\ref{tab:audio_correlation}, we report the accuracy with respect to the ground-truth gender labels of low-resolution images and audios.\\
\noindent\hspace{-0.7em}\textbf{Failure Cases.} We observe that failures may correspond more to the inherent bias presented in the training set than the training algorithm or network architecture. Failure cases sometimes happen when the gender can be easily guessed just from the low-resolution image. Some of the failure cases are reported in Fig~\ref{fig:failure}.

\section{Conclusions}
We have introduced a new paradigm for face super-resolution, where also audio contributes to the restoration of missing details in the low-resolution input image. We have described the design of a neural network and the corresponding training procedure to successfully make use of the audio signal despite the difficulty of extracting visual information from it. We have also shown quantitatively that audio can contribute to improving the accuracy of the identity as well as the gender and age of the restored face. Moreover, we have shown that it is possible to mix low-resolution images and audios from different videos and obtain semantically meaningful high resolution images. A fundamental challenge in this work was the fusion of information from these very different modalities. As we have shown, valuable information is present in both. However, we observed that a na\"ive end-to-end training tends to ignore audio information. We conjecture that this problem might be a fundamental issue with current training schemes of neural networks and its solution could provide insights on how to improve the training on tasks in general. 

\noindent \textbf{Acknowledgements.} This work was supported by grants 165845\&169622 of the Swiss National Science Foundation.

{\small
\bibliographystyle{ieee_fullname}
\bibliography{egbib}

\begin{thebibliography}{10}\itemsep=-1pt

\bibitem{Ahn_2018_ECCV}
Namhyuk Ahn, Byungkon Kang, and Kyung-Ah Sohn.
\newblock Fast, accurate, and lightweight super-resolution with cascading
  residual network.
\newblock In {\em The European Conference on Computer Vision (ECCV)}, September
  2018.

\bibitem{Arandjelovic_2018_ECCV}
Relja Arandjelovic and Andrew Zisserman.
\newblock Objects that sound.
\newblock In {\em The European Conference on Computer Vision (ECCV)}, September
  2018.

\bibitem{Bau_2019_ICCV}
David Bau, Jun-Yan Zhu, Jonas Wulff, William Peebles, Hendrik Strobelt, Bolei
  Zhou, and Antonio Torralba.
\newblock Seeing what a gan cannot generate.
\newblock In {\em The IEEE International Conference on Computer Vision (ICCV)},
  October 2019.

\bibitem{Bulat_2018_CVPR}
Adrian Bulat and Georgios Tzimiropoulos.
\newblock Super-fan: Integrated facial landmark localization and
  super-resolution of real-world low resolution faces in arbitrary poses with
  gans.
\newblock In {\em The IEEE Conference on Computer Vision and Pattern
  Recognition (CVPR)}, June 2018.

\bibitem{Bulat_2018_ECCV}
Adrian Bulat, Jing Yang, and Georgios Tzimiropoulos.
\newblock To learn image super-resolution, use a gan to learn how to do image
  degradation first.
\newblock In {\em The European Conference on Computer Vision (ECCV)}, September
  2018.

\bibitem{Cai_2019_ICCV}
Jianrui Cai, Hui Zeng, Hongwei Yong, Zisheng Cao, and Lei Zhang.
\newblock Toward real-world single image super-resolution: A new benchmark and
  a new model.
\newblock In {\em The IEEE International Conference on Computer Vision (ICCV)},
  October 2019.

\bibitem{Chen_2019_CVPR}
Chang Chen, Zhiwei Xiong, Xinmei Tian, Zheng-Jun Zha, and Feng Wu.
\newblock Camera lens super-resolution.
\newblock In {\em The IEEE Conference on Computer Vision and Pattern
  Recognition (CVPR)}, June 2019.

\bibitem{Chen_2018_CVPR}
Yu Chen, Ying Tai, Xiaoming Liu, Chunhua Shen, and Jian Yang.
\newblock Fsrnet: End-to-end learning face super-resolution with facial priors.
\newblock In {\em The IEEE Conference on Computer Vision and Pattern
  Recognition (CVPR)}, June 2018.

\bibitem{Choi_2019_ICCV}
Jun-Ho Choi, Huan Zhang, Jun-Hyuk Kim, Cho-Jui Hsieh, and Jong-Seok Lee.
\newblock Evaluating robustness of deep image super-resolution against
  adversarial attacks.
\newblock In {\em The IEEE International Conference on Computer Vision (ICCV)},
  October 2019.

\bibitem{Chung18b}
J.~S. Chung, A. Nagrani, and A. Zisserman.
\newblock Voxceleb2: Deep speaker recognition.
\newblock In {\em INTERSPEECH}, 2018.

\bibitem{Dai_2019_CVPR}
Tao Dai, Jianrui Cai, Yongbing Zhang, Shu-Tao Xia, and Lei Zhang.
\newblock Second-order attention network for single image super-resolution.
\newblock In {\em The IEEE Conference on Computer Vision and Pattern
  Recognition (CVPR)}, June 2019.

\bibitem{Deng_2019_ICCV}
Xin Deng, Ren Yang, Mai Xu, and Pier~Luigi Dragotti.
\newblock Wavelet domain style transfer for an effective perception-distortion
  tradeoff in single image super-resolution.
\newblock In {\em The IEEE International Conference on Computer Vision (ICCV)},
  October 2019.

\bibitem{ephrat2018looking}
A. Ephrat, I. Mosseri, O. Lang, T. Dekel, K Wilson, A. Hassidim, W.~T. Freeman,
  and M. Rubinstein.
\newblock Looking to listen at the cocktail party: A speaker-independent
  audio-visual model for speech separation.
\newblock {\em arXiv preprint arXiv:1804.03619}, 2018.

\bibitem{Gao_2018_ECCV}
Ruohan Gao, Rogerio Feris, and Kristen Grauman.
\newblock Learning to separate object sounds by watching unlabeled video.
\newblock In {\em The European Conference on Computer Vision (ECCV)}, September
  2018.

\bibitem{NIPS2014_Goodfellow}
Ian Goodfellow, Jean Pouget-Abadie, Mehdi Mirza, Bing Xu, David Warde-Farley,
  Sherjil Ozair, Aaron Courville, and Yoshua Bengio.
\newblock Generative adversarial nets.
\newblock In Z. Ghahramani, M. Welling, C. Cortes, N.~D. Lawrence, and K.~Q.
  Weinberger, editors, {\em Advances in Neural Information Processing Systems
  27}, pages 2672--2680. Curran Associates, Inc., 2014.

\bibitem{Gu_2019_CVPR}
Jinjin Gu, Hannan Lu, Wangmeng Zuo, and Chao Dong.
\newblock Blind super-resolution with iterative kernel correction.
\newblock In {\em The IEEE Conference on Computer Vision and Pattern
  Recognition (CVPR)}, June 2019.

\bibitem{Han_2018_CVPR}
Wei Han, Shiyu Chang, Ding Liu, Mo Yu, Michael Witbrock, and Thomas~S. Huang.
\newblock Image super-resolution via dual-state recurrent networks.
\newblock In {\em The IEEE Conference on Computer Vision and Pattern
  Recognition (CVPR)}, June 2018.

\bibitem{Haris_2018_CVPR}
Muhammad Haris, Gregory Shakhnarovich, and Norimichi Ukita.
\newblock Deep back-projection networks for super-resolution.
\newblock In {\em The IEEE Conference on Computer Vision and Pattern
  Recognition (CVPR)}, June 2018.

\bibitem{Haris_2019_CVPR}
Muhammad Haris, Gregory Shakhnarovich, and Norimichi Ukita.
\newblock Recurrent back-projection network for video super-resolution.
\newblock In {\em The IEEE Conference on Computer Vision and Pattern
  Recognition (CVPR)}, June 2019.

\bibitem{He_2019_CVPR}
Xiangyu He, Zitao Mo, Peisong Wang, Yang Liu, Mingyuan Yang, and Jian Cheng.
\newblock Ode-inspired network design for single image super-resolution.
\newblock In {\em The IEEE Conference on Computer Vision and Pattern
  Recognition (CVPR)}, June 2019.

\bibitem{Hu_2019_CVPR}
Xuecai Hu, Haoyuan Mu, Xiangyu Zhang, Zilei Wang, Tieniu Tan, and Jian Sun.
\newblock Meta-sr: A magnification-arbitrary network for super-resolution.
\newblock In {\em The IEEE Conference on Computer Vision and Pattern
  Recognition (CVPR)}, June 2019.

\bibitem{Huang_2017_ICCV}
Huaibo Huang, Ran He, Zhenan Sun, and Tieniu Tan.
\newblock Wavelet-srnet: A wavelet-based cnn for multi-scale face super
  resolution.
\newblock In {\em The IEEE International Conference on Computer Vision (ICCV)},
  Oct 2017.

\bibitem{Hui_2018_CVPR}
Zheng Hui, Xiumei Wang, and Xinbo Gao.
\newblock Fast and accurate single image super-resolution via information
  distillation network.
\newblock In {\em The IEEE Conference on Computer Vision and Pattern
  Recognition (CVPR)}, June 2018.

\bibitem{karras2018style}
Tero Karras, Samuli Laine, and Timo Aila.
\newblock A style-based generator architecture for generative adversarial
  networks.
\newblock {\em arXiv preprint arXiv:1812.04948}, 2018.

\bibitem{Kim_2019_ICCV}
Soo~Ye Kim, Jihyong Oh, and Munchurl Kim.
\newblock Deep sr-itm: Joint learning of super-resolution and inverse
  tone-mapping for 4k uhd hdr applications.
\newblock In {\em The IEEE International Conference on Computer Vision (ICCV)},
  October 2019.

\bibitem{kingma2014adam}
Diederik~P Kingma and Jimmy Ba.
\newblock Adam: A method for stochastic optimization.
\newblock {\em arXiv preprint arXiv:1412.6980}, 2014.

\bibitem{Lai_2017_CVPR}
Wei-Sheng Lai, Jia-Bin Huang, Narendra Ahuja, and Ming-Hsuan Yang.
\newblock Deep laplacian pyramid networks for fast and accurate
  super-resolution.
\newblock In {\em The IEEE Conference on Computer Vision and Pattern
  Recognition (CVPR)}, July 2017.

\bibitem{Ledig_2017_CVPR}
Christian Ledig, Lucas Theis, Ferenc Huszar, Jose Caballero, Andrew Cunningham,
  Alejandro Acosta, Andrew Aitken, Alykhan Tejani, Johannes Totz, Zehan Wang,
  and Wenzhe Shi.
\newblock Photo-realistic single image super-resolution using a generative
  adversarial network.
\newblock In {\em The IEEE Conference on Computer Vision and Pattern
  Recognition (CVPR)}, July 2017.

\bibitem{Li_2018_ECCV}
Juncheng Li, Faming Fang, Kangfu Mei, and Guixu Zhang.
\newblock Multi-scale residual network for image super-resolution.
\newblock In {\em The European Conference on Computer Vision (ECCV)}, September
  2018.

\bibitem{Li_2019_CVPR_1}
Sheng Li, Fengxiang He, Bo Du, Lefei Zhang, Yonghao Xu, and Dacheng Tao.
\newblock Fast spatio-temporal residual network for video super-resolution.
\newblock In {\em The IEEE Conference on Computer Vision and Pattern
  Recognition (CVPR)}, June 2019.

\bibitem{Li_2019_CVPR}
Zhen Li, Jinglei Yang, Zheng Liu, Xiaomin Yang, Gwanggil Jeon, and Wei Wu.
\newblock Feedback network for image super-resolution.
\newblock In {\em The IEEE Conference on Computer Vision and Pattern
  Recognition (CVPR)}, June 2019.

\bibitem{oh2019speech2face}
T.~H. Oh, T. Dekel, C. Kim, I. Mosseri, W.~T. Freeman, M. Rubinstein, and W.
  Matusik.
\newblock Speech2face: Learning the face behind a voice.
\newblock In {\em CVPR}, 2019.

\bibitem{Owens_2018_ECCV}
Andrew Owens and Alexei~A. Efros.
\newblock Audio-visual scene analysis with self-supervised multisensory
  features.
\newblock In {\em The European Conference on Computer Vision (ECCV)}, September
  2018.

\bibitem{Park_2018_ECCV}
Seong-Jin Park, Hyeongseok Son, Sunghyun Cho, Ki-Sang Hong, and Seungyong Lee.
\newblock Srfeat: Single image super-resolution with feature discrimination.
\newblock In {\em The European Conference on Computer Vision (ECCV)}, September
  2018.

\bibitem{Parkhi15}
O.~M. Parkhi, A. Vedaldi, and A. Zisserman.
\newblock Deep face recognition.
\newblock In {\em British Machine Vision Conference}, 2015.

\bibitem{Qiu_2019_ICCV}
Yajun Qiu, Ruxin Wang, Dapeng Tao, and Jun Cheng.
\newblock Embedded block residual network: A recursive restoration model for
  single-image super-resolution.
\newblock In {\em The IEEE International Conference on Computer Vision (ICCV)},
  October 2019.

\bibitem{Rad_2019_ICCV}
Mohammad~Saeed Rad, Behzad Bozorgtabar, Urs-Viktor Marti, Max Basler,
  Hazim~Kemal Ekenel, and Jean-Philippe Thiran.
\newblock Srobb: Targeted perceptual loss for single image super-resolution.
\newblock In {\em The IEEE International Conference on Computer Vision (ICCV)},
  October 2019.

\bibitem{Rothe2018}
Rasmus Rothe, Radu Timofte, and Luc Van~Gool.
\newblock Deep expectation of real and apparent age from a single image without
  facial landmarks.
\newblock {\em International Journal of Computer Vision}, 126(2):144--157, Apr
  2018.

\bibitem{Shlizerman_2018_CVPR}
Eli Shlizerman, Lucio Dery, Hayden Schoen, and Ira Kemelmacher-Shlizerman.
\newblock Audio to body dynamics.
\newblock In {\em The IEEE Conference on Computer Vision and Pattern
  Recognition (CVPR)}, June 2018.

\bibitem{Soh_2019_CVPR}
Jae~Woong Soh, Gu~Yong Park, Junho Jo, and Nam~Ik Cho.
\newblock Natural and realistic single image super-resolution with explicit
  natural manifold discrimination.
\newblock In {\em The IEEE Conference on Computer Vision and Pattern
  Recognition (CVPR)}, June 2019.

\bibitem{song2018talking}
Yang Song, Jingwen Zhu, Xiaolong Wang, and Hairong Qi.
\newblock Talking face generation by conditional recurrent adversarial network.
\newblock {\em arXiv preprint arXiv:1804.04786}, 2018.

\bibitem{Sterling_2018_ECCV}
Auston Sterling, Justin Wilson, Sam Lowe, and Ming~C. Lin.
\newblock Isnn: Impact sound neural network for audio-visual object
  classification.
\newblock In {\em The European Conference on Computer Vision (ECCV)}, September
  2018.

\bibitem{Tai_2017_CVPR}
Ying Tai, Jian Yang, and Xiaoming Liu.
\newblock Image super-resolution via deep recursive residual network.
\newblock In {\em The IEEE Conference on Computer Vision and Pattern
  Recognition (CVPR)}, July 2017.

\bibitem{Tian_2018_ECCV}
Yapeng Tian, Jing Shi, Bochen Li, Zhiyao Duan, and Chenliang Xu.
\newblock Audio-visual event localization in unconstrained videos.
\newblock In {\em The European Conference on Computer Vision (ECCV)}, September
  2018.

\bibitem{Wang_2019_CVPR}
Longguang Wang, Yingqian Wang, Zhengfa Liang, Zaiping Lin, Jungang Yang, Wei
  An, and Yulan Guo.
\newblock Learning parallax attention for stereo image super-resolution.
\newblock In {\em The IEEE Conference on Computer Vision and Pattern
  Recognition (CVPR)}, June 2019.

\bibitem{DBLP:journals/corr/abs-1905-12681}
Weiyao Wang, Du Tran, and Matt Feiszli.
\newblock What makes training multi-modal networks hard?
\newblock {\em CoRR}, abs/1905.12681, 2019.

\bibitem{Wang_2018_CVPR}
Xintao Wang, Ke Yu, Chao Dong, and Chen Change~Loy.
\newblock Recovering realistic texture in image super-resolution by deep
  spatial feature transform.
\newblock In {\em The IEEE Conference on Computer Vision and Pattern
  Recognition (CVPR)}, June 2018.

\bibitem{Xu_2019_CVPR}
Xiangyu Xu, Yongrui Ma, and Wenxiu Sun.
\newblock Towards real scene super-resolution with raw images.
\newblock In {\em The IEEE Conference on Computer Vision and Pattern
  Recognition (CVPR)}, June 2019.

\bibitem{Xu_2017_ICCV}
Xiangyu Xu, Deqing Sun, Jinshan Pan, Yujin Zhang, Hanspeter Pfister, and
  Ming-Hsuan Yang.
\newblock Learning to super-resolve blurry face and text images.
\newblock In {\em The IEEE International Conference on Computer Vision (ICCV)},
  Oct 2017.

\bibitem{Yi_2019_ICCV}
Peng Yi, Zhongyuan Wang, Kui Jiang, Junjun Jiang, and Jiayi Ma.
\newblock Progressive fusion video super-resolution network via exploiting
  non-local spatio-temporal correlations.
\newblock In {\em The IEEE International Conference on Computer Vision (ICCV)},
  October 2019.

\bibitem{Yu_2018_ECCV}
Xin Yu, Basura Fernando, Bernard Ghanem, Fatih Porikli, and Richard Hartley.
\newblock Face super-resolution guided by facial component heatmaps.
\newblock In {\em The European Conference on Computer Vision (ECCV)}, September
  2018.

\bibitem{Yu_2018_CVPR}
Xin Yu, Basura Fernando, Richard Hartley, and Fatih Porikli.
\newblock Super-resolving very low-resolution face images with supplementary
  attributes.
\newblock In {\em The IEEE Conference on Computer Vision and Pattern
  Recognition (CVPR)}, June 2018.

\bibitem{Yu_2017_CVPR}
Xin Yu and Fatih Porikli.
\newblock Hallucinating very low-resolution unaligned and noisy face images by
  transformative discriminative autoencoders.
\newblock In {\em The IEEE Conference on Computer Vision and Pattern
  Recognition (CVPR)}, July 2017.

\bibitem{Zhang_2019_ICCV}
Haochen Zhang, Dong Liu, and Zhiwei Xiong.
\newblock Two-stream action recognition-oriented video super-resolution.
\newblock In {\em The IEEE International Conference on Computer Vision (ICCV)},
  October 2019.

\bibitem{Zhang_2018_ECCV_1}
Kaipeng Zhang, Zhanpeng Zhang, Chia-Wen Cheng, Winston~H. Hsu, Yu Qiao, Wei
  Liu, and Tong Zhang.
\newblock Super-identity convolutional neural network for face hallucination.
\newblock In {\em The European Conference on Computer Vision (ECCV)}, September
  2018.

\bibitem{Zhang_2018_CVPRR}
Kai Zhang, Wangmeng Zuo, and Lei Zhang.
\newblock Learning a single convolutional super-resolution network for multiple
  degradations.
\newblock In {\em The IEEE Conference on Computer Vision and Pattern
  Recognition (CVPR)}, June 2018.

\bibitem{Zhang_2019_ICCV_1}
Wenlong Zhang, Yihao Liu, Chao Dong, and Yu Qiao.
\newblock Ranksrgan: Generative adversarial networks with ranker for image
  super-resolution.
\newblock In {\em The IEEE International Conference on Computer Vision (ICCV)},
  October 2019.

\bibitem{Zhang_2019_CVPR}
Xuaner Zhang, Qifeng Chen, Ren Ng, and Vladlen Koltun.
\newblock Zoom to learn, learn to zoom.
\newblock In {\em The IEEE Conference on Computer Vision and Pattern
  Recognition (CVPR)}, June 2019.

\bibitem{Zhang_2018_ECCV}
Yulun Zhang, Kunpeng Li, Kai Li, Lichen Wang, Bineng Zhong, and Yun Fu.
\newblock Image super-resolution using very deep residual channel attention
  networks.
\newblock In {\em The European Conference on Computer Vision (ECCV)}, September
  2018.

\bibitem{Zhang_2018_CVPR}
Yulun Zhang, Yapeng Tian, Yu Kong, Bineng Zhong, and Yun Fu.
\newblock Residual dense network for image super-resolution.
\newblock In {\em The IEEE Conference on Computer Vision and Pattern
  Recognition (CVPR)}, June 2018.

\bibitem{Zhang_2019_CVPR_1}
Zhifei Zhang, Zhaowen Wang, Zhe Lin, and Hairong Qi.
\newblock Image super-resolution by neural texture transfer.
\newblock In {\em The IEEE Conference on Computer Vision and Pattern
  Recognition (CVPR)}, June 2019.

\bibitem{Zhao_2018_ECCV}
Hang Zhao, Chuang Gan, Andrew Rouditchenko, Carl Vondrick, Josh McDermott, and
  Antonio Torralba.
\newblock The sound of pixels.
\newblock In {\em The European Conference on Computer Vision (ECCV)}, September
  2018.

\bibitem{Zhou_2019_ICCV}
Ruofan Zhou and Sabine Susstrunk.
\newblock Kernel modeling super-resolution on real low-resolution images.
\newblock In {\em The IEEE International Conference on Computer Vision (ICCV)},
  October 2019.

\bibitem{zhu2018high}
Hao Zhu, Aihua Zheng, Huaibo Huang, and Ran He.
\newblock High-resolution talking face generation via mutual information
  approximation.
\newblock {\em arXiv preprint arXiv:1812.06589}, 2018.

\end{thebibliography}
}

\end{document}